\documentclass[conference]{IEEEtran}
\IEEEoverridecommandlockouts

\usepackage{cite}
\usepackage{amsmath,amssymb,amsfonts}
\usepackage{algorithmic}
\usepackage{graphicx}
\usepackage{textcomp}
\usepackage{xcolor}
\usepackage{multirow}
\usepackage{hyperref}
\usepackage{float}
\usepackage{booktabs}
\usepackage[numbers]{natbib}
\usepackage{soul}

\def\BibTeX{{\rm B\kern-.05em{\sc i\kern-.025em b}\kern-.08em
    T\kern-.1667em\lower.7ex\hbox{E}\kern-.125emX}}
\begin{document}
\makeatletter
\newcommand{\linebreakand}{%
  \end{@IEEEauthorhalign}
  \hfill\mbox{}\par
  \mbox{}\hfill\begin{@IEEEauthorhalign}
}
\makeatother
\title{MixMAS: A Framework for Sampling-Based Mixer Architecture Search for Multimodal Fusion and Learning
}


\author{\IEEEauthorblockN{Abdelmadjid Chergui}
\IEEEauthorblockA{\textit{Higher School of Computer Science} \\
\textit{8 Mai 1945}\\
SBA, Algeria \\
a.chergui@esi-sba.dz}
\and
\IEEEauthorblockN{Grigor Bezirganyan}
\IEEEauthorblockA{\textit{Aix-Marseille Univ, LIS, CNRS} \\
Marseille, France \\
grigor.bezirganyan@univ-amu.fr}
\and
\IEEEauthorblockN{Sana Sellami}
\IEEEauthorblockA{\textit{Aix-Marseille Univ, LIS, CNRS} \\
Marseille, France \\
sana.sellami@univ-amu.fr}
\linebreakand
\IEEEauthorblockN{Laure Berti-Équille}
\IEEEauthorblockA{\textit{IRD, ESPACE-DEV} \\
Montpellier, France \\
laure.berti@ird.fr}
\and
\IEEEauthorblockN{Sébastien Fournier}
\IEEEauthorblockA{\textit{Aix-Marseille Univ, LIS, CNRS} \\
Marseille, France \\
sebastien.fournier@univ-amu.fr}
}

\maketitle

\begin{abstract}
Choosing a suitable deep learning architecture for multimodal data fusion is a challenging task, as it requires the effective integration and processing of diverse data types, each with distinct structures and characteristics. In this paper, we introduce MixMAS, a novel framework for sampling-based mixer architecture search tailored to multimodal learning. Our approach automatically selects the optimal MLP-based architecture for a given multimodal machine learning (MML) task. Specifically, MixMAS utilizes a sampling-based micro-benchmarking strategy to explore various combinations of modality-specific encoders, fusion functions, and fusion networks, systematically identifying the architecture that best meets the task’s performance metrics. 

\end{abstract}


\section{Introduction}\label{sec:introduction}

The increasing complexity and diversity of data in various domains require the use of multimodal learning, which can leverage and integrate information from different modalities including text, image, audio, video, time series, etc. \cite{baltrusaitis2017multimodal}.
The application of multimodal learning spans a wide array of fields, including but not limited to, text-to-image generation, text-to-video synthesis, robotics, and autonomous driving \cite{liang2022foundations}.
The essence of multimodal learning lies in its ability
to provide a more holistic understanding of the data by harnessing the complementary nature of different data types.
However, the fusion of multimodal data presents significant computational and theoretical challenges \cite{liang2022foundations}
that arise from the inherent heterogeneity of the data sources, which makes learning inter-modal relationships and representations more difficult, as each modality often exhibits diverse qualities, structures, and relevance to the task at hand. Computationally, processing and fusing such diverse data types at scale is very hardware demanding. This highlights the increasing need for specialized architectures that can effectively handle multimodal data.

In response to these challenges,
Multi-Layer Perceptron (MLP) based architectures have emerged as a promising solution \cite{tolstikhin2021, mai2023hypermixer, osti_10494106}.
These architectures offer a compelling alternative to transformer models, as achieve a favorable balance between performance and computational complexity \cite{tolstikhin2021}.
The advantages of these architectures lie in being computationally more efficient \cite{tolstikhin2021},
exhibiting a simpler architectural design \cite{liu2022ready}, facilitating ease of implementation and modification,
and robustness in handling a variety of data types and tasks \cite{liu2022ready}.
Integrating these components into an automated search pipeline enables us to leverage their benefits and design architectures that address the unique requirements and constraints of each task, resulting in more effective and specialized models.

In this paper, we present MixMAS, an automated framework for solving the problem of choosing which MLP-based architecture to use for a given multimodal machine learning task (MML).
Our contributions can be summarized as follows: 1) We propose an automated pipeline for selecting the optimal MLP-based architecture for multimodal tasks. This pipeline benchmarks various MLP-based encoders for each modality, identifies the most effective fusion function for integrating all modalities, and determines the most suitable fusion network. Although we selected MLP-based architectures for their computational and conceptual simplicity, the framework is flexible and can accommodate other models, such as transformers, CNNs, and more. 
2) We propose to employ a sampling approach, where different modules are benchmarked only on a small sample of the dataset to reduce the computational cost compared to evaluating on the full dataset. 
3) We experimentally validate that our proposed pipeline for optimizing multimodal MLP-based architectures improves accuracy compared to standard MLP-based multimodal networks.
4) We open-source the code for the proposed framework at \href{https://github.com/Madjid-CH/auto-mixer}{https://github.com/Madjid-CH/auto-mixer}.

\section{Related Work}\label{sec:related-work}
This section reviews the related research on MLP-based models and multimodal architecture search methods.

\textbf {MLP-Mixers}~\cite{tolstikhin2021} introduced a paradigm shift in the field of deep learning
by achieving competitive results on image classification benchmarks against the state-of-the-art models with comparable computational resources.
The architecture is based exclusively on MLPs with two types of layers:
MLPs applied independently to image patches, and MLPs applied across patches.
Many follow-up works improve the MLP-Mixer architecture, such as: Region-aware MLP (\textbf{RaMLP})~\cite{LaiDGZ23}, which addresses the limitation of fixed input sizes in previous MLP models and captures both local and global visual cues in a region-aware manner; the
\textbf{HyperMixer}~\cite{mai2023hypermixer} introduces a token mixing mechanism called HyperMixing, which uses hypernetworks to dynamically generate the weights of the token mixing MLP based on the input. This allows HyperMixer to handle variable input lengths and ensures systematicity by modeling interactions between tokens with shared weights across positions; the \textbf{Monarch-Mixer}~\cite{osti_10494106} uses Monarch matrices for efficient performance on GPUs and demonstrates comparable or superior results in tasks like language modeling or image classification, with fewer parameters.
These models are conceptually simpler compared to other architectures like CNNs and Transformers.
They also make a good trade-off between performance and computational efficiency.

The M2-Mixer \cite{bezirganyan2023} architecture has been proposed for multimodal classification, leveraging the simplicity and efficiency of MLP-Mixers. It employs a multi-head loss function to address optimization imbalance, ensuring that no single modality dominates the learning process. This results in a conceptually and computationally simple model that outperforms baseline models on benchmark multimodal datasets, achieving higher accuracy and significantly reducing training time. However, MLP-Mixers are not universally effective across all modalities, and selecting the right MLP-based architecture to optimize performance can be challenging. To address this, our proposed pipeline systematically identifies the most suitable MLP-based network for each specific dataset, ensuring optimal performance and compatibility with the data characteristics.

\textbf{Multimodal NAS:} Neural Architecture Search (NAS) methods try to automate the process of finding the optimal architecture for given tasks and datasets. Several multimodal NAS approaches were presented in the literature using different search algorithm and search space \cite{perez2019mfas, yu2020deep, yin2022bm, xu2021mufasa}. These approaches are often complex to implement and train, and adjusting certain parts of the architecture typically requires rerunning the entire NAS process. In contrast, our proposed framework is simple to implement and allows for the retention of existing micro-benchmarks, enabling quick updates to the architecture when new modules or components are added to the search space. Additionally, the modular design of the pipeline permits micro-benchmarking to be applied selectively to specific parts of the architecture, streamlining the process.

\section{Our  Framework}\label{sec:method}

We propose \textbf{MixMAS}, a sampling-based framework for automatic mixer architecture search in multimodal learning. MixMAS efficiently selects optimal MLP-based architectures for various multimodal tasks, leveraging modularity and extensibility for adaptability. This pipeline focuses on efficient MLP architectures composed solely of matrix multiplications, transformations, and activation functions, streamlining the search process for suitable configurations in multimodal learning.

As illustrated in Figure \ref{fig:generic_architecture}, our pipeline can be conceptually divided into four main stages: 1) Sampling, 2) Encoder selection,  3) Fusion function selection, 4) Fusion Network selection.

\begin{figure}[h]
    \centering
    \includegraphics[width=1\linewidth]{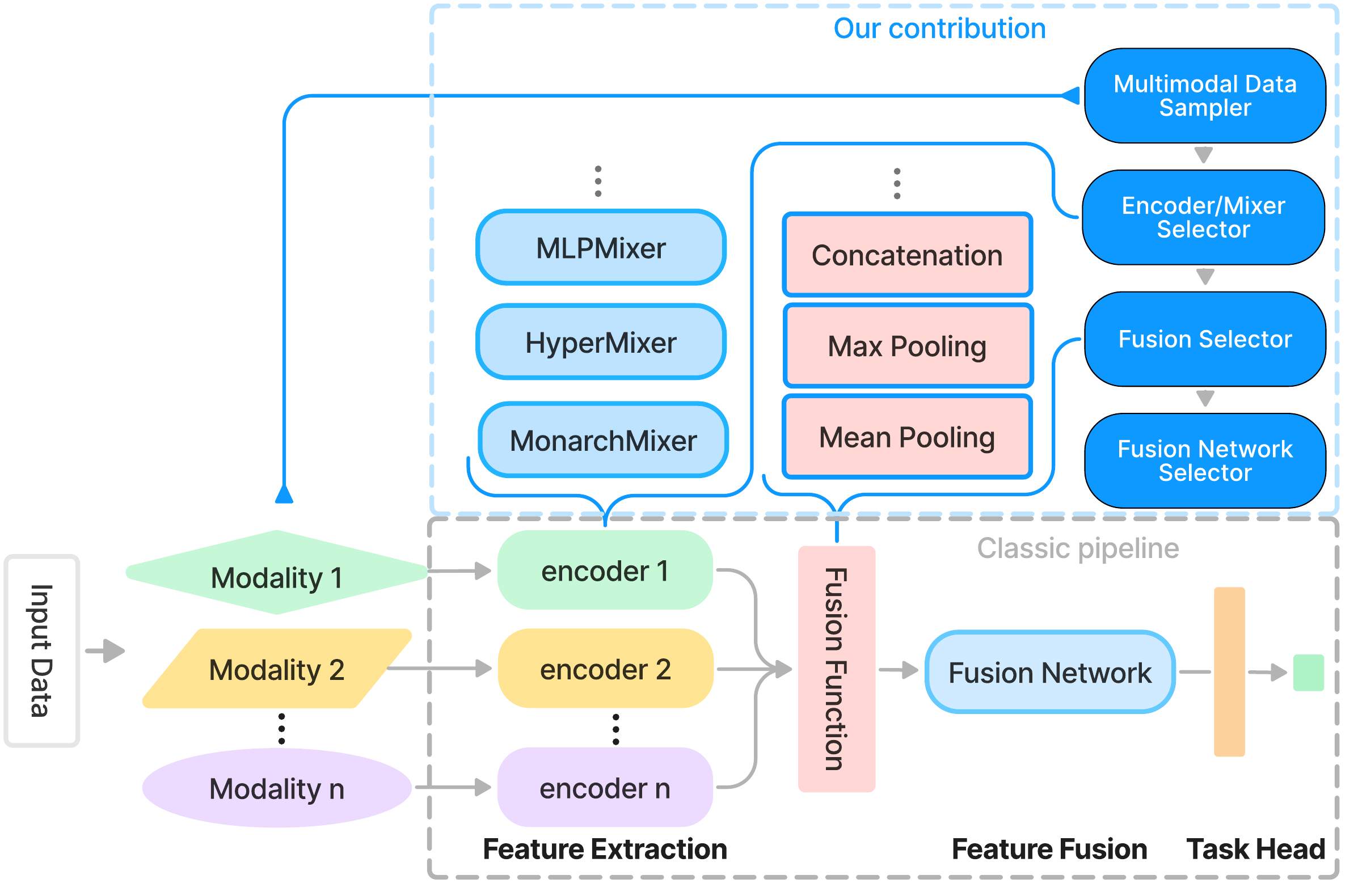}
    \caption{MixMAS sampling based architecture search pipeline}
    \label{fig:generic_architecture}
\end{figure}
\textbf{Sampling:} 
The sampling module selects a subset of the dataset for comparing the performance of different modules at each stage. It ensures that the sampled subset is representative of the entire dataset, which is essential for obtaining accurate and reliable performance metrics to guide the selection process.
The number of samples is computed as in \eqref{eq:sampling} using the sample size determination formula \cite{hogg2013probability}:
\begin{gather}
    N^{\prime} = \frac{n}{1+\frac{z^2 \times \hat{p}(1-\hat{p})}{\varepsilon^2 N}}, \quad
    n = \frac{z^2 \times \hat{p}(1-\hat{p})}{\varepsilon^2},
    \label{eq:sampling}
\end{gather}
where $N$ is the size of the dataset, $z$ is the z-score, $\hat{p}$ is the estimated proportion of the population that has the attribute of interest, and $\varepsilon$ is the margin of error (i.e., 1\%). We use random sampling to approximate the class distribution of the original dataset. To validate this approach, we calculate the distance between the class proportions in the original and sampled datasets, ensuring it remains below 0.05. In future work, we aim to implement uncertainty-based sampling to obtain more informative samples.

\textbf{Encoder Selection} involves benchmarking the performance of various MLP-based encoders on the sampled dataset for each modality. The choice of encoders depends on the dataset's modalities, ensuring all options are MLP-based. Users can customize the evaluation metrics within the framework to suit the problem's specific nature. For example, in a balanced binary classification problem, accuracy might be preferred, while for imbalanced classes, metrics like precision, recall, or F1 score may be more appropriate.
While MLP based encoders were selected for this work, the framework allows to incorporate arbitrary encoders, such as transformers, CNNs, RNNs, etc.
The best encoder for each modality is then selected based on the benchmarking results to be used in the next stage.

\textbf{Fusion Function Selection} entails choosing the best fusion function to combine features from each modality. We use intermediate fusion, as raw fusion is often impractical with differing data structures, and late fusion may be suboptimal for highly correlated modalities \cite{stahlschmidt2022multimodal}. The fusion function is evaluated based on classification performance.

\textbf{Fusion Network Selection} is responsible for selecting the appropriate network for encoding cross-modal information and preparing the last embedding for the task head.
Similar to the Encoder Selection, multiple MLP-based networks will be benchmarked. The encoders for each modality and the fusion function are the ones fixed from the previous stages. 

A recommended approach is to retain the micro-benchmarking scores after identifying the final architecture. By avoiding the need to re-benchmark previously evaluated architectures, this strategy streamlines the process when incorporating new components or making adjustments.

\section{Experiments}\label{sec:experiments}
\begin{table*}[ht]
    \caption{Micro Benchmark Results for Each Stage. The module with the highest score will be selected.}
    \label{tab:micro_benchmark}
    \centering
    \begin{tabular}{lcccccccccccccccc}
        \toprule

        \multicolumn{5}{c}{\textbf{MM-IMDB}} & \multicolumn{5}{c}{\textbf{AV-MNIST}} & \multicolumn{5}{c}{\textbf{MIMIC-III}} \\
        \midrule
        Sampling(\%) & \multicolumn{3}{c}{23\%} && \multicolumn{4}{c}{12\%} && \multicolumn{5}{c}{21\%} \\
        \midrule
        \midrule
        Module  & Score F1-w(\%) & & & & & Module  & Score Acc(\%) & & & & & Module  & Score Acc(\%) \\
        \midrule
        \multicolumn{5}{c}{\textbf{Image Encoder Selection}} & \multicolumn{5}{c}{\textbf{Image Encoder Selection}} & \multicolumn{5}{c}{\textbf{Time-Series Encoder Selection}} \\
        \midrule
        \textbf{MLPMixer} & $\mathbf{24.02}$ & & & & & MLPMixer  & $44.27$ & & & & & MLPMixer  & $40.77$ \\
        HyperMixer & $16.89$ & & & & & \textbf{HyperMixer}  & $\mathbf{56.15}$ & & & & & \textbf{HyperMixer}  & $\mathbf{45.36}$ \\
        RaMLP & $14.44$ & & & & & RaMLP  & $47.52$ & & & & & MonarchMixer  & $44.38$ \\
        \midrule[1pt]
        \multicolumn{5}{c}{\textbf{Text Encoder Selection}} & \multicolumn{5}{c}{\textbf{Audio Encoder Selection}} & \multicolumn{5}{c}{\textbf{Tabular Encoder Selection}} \\
        \midrule
        MLPMixer & $9.20$ & & & & & MLPMixer  & $27.40$ & & & & & ---  & --- \\
        HyperMixer & $15.07$ & & & & & \textbf{HyperMixer}  & $\mathbf{29.16}$ & & & & & ---  & --- \\
        \textbf{MonarchMixer} & $\mathbf{28.55}$ & & & & & MonarchMixer  & $28.49$ & & & & & ---  & --- \\
        \midrule[1pt]
        \multicolumn{5}{c}{\textbf{Fusion Function Selection}} & \multicolumn{5}{c}{\textbf{Fusion Function Selection}} & \multicolumn{5}{c}{\textbf{Fusion Function Selection}} \\
        \midrule
        \textbf{ConcatFusion} & $\mathbf{19.56}$ & & & & & \textbf{ConcatFusion}  & $\mathbf{18.38}$ & & & & & \textbf{ConcatFusion}  & $\mathbf{28.55}$ \\
        MeanFusion & $10.20$ & & & & & MeanFusion  & $9.61$ & & & & & MeanFusion  & $4.28$ \\
        MaxFusion & $9.07$ & & & & & MaxFusion  & $6.20$ & & & & & MaxFusion  & $6.73$ \\
        \midrule[1pt]
        \multicolumn{5}{c}{\textbf{Fusion Network Selection}} & \multicolumn{5}{c}{\textbf{Fusion Network Selection}} & \multicolumn{5}{c}{\textbf{Fusion Network Selection}} \\
        \midrule
        \textbf{HyperMixer} & $\mathbf{29.0}$ & & & & & \textbf{HyperMixer}  & $\mathbf{53.47}$ & & & & & \textbf{HyperMixer}  & $\mathbf{38.15}$ \\
        MLPMixer & $25.97$ & & & & & MLPMixer  & $42.17$ & & & & & MLPMixer  & $34.14$ \\
        \bottomrule
    \end{tabular}
\end{table*}

\begin{table*}[ht]
    \caption{Results on \textbf{MM-IMDB}, \textbf{AV-MNIST} and \textbf{MIMIC-III} datasets.}
    \label{tab:merged_res}
    \centering
    \small
    \begin{tabular}{lrrrrrr}
        \toprule 
        & \multicolumn{2}{c}{\textbf{MM-IMDB}} & \multicolumn{2}{c}{\textbf{AV-MNIST}} & \multicolumn{2}{c}{\textbf{MIMIC-III}} \\
        \cmidrule(lr){2-7} 
        Architecture & $\begin{array}{r}
                                     \text{F1-w. (\%)} \\
                                     \text {(avg)}
        \end{array}$
        & $\begin{array}{r}
               \text{Training} \\
               \text {Params (M)}
        \end{array}$ & $\begin{array}{r}
                                     \text{Acc. (\%)} \\
                                     \text {(avg)}
        \end{array}$
        & $\begin{array}{r}
               \text{Training} \\
               \text {Params (M)}
        \end{array}$ & $\begin{array}{r}
                                     \text{Acc. (\%)} \\
                                     \text {(avg)}
        \end{array}$ 
        & $\begin{array}{r}
               \text{Training} \\
               \text {Params (M)}
        \end{array}$ \\
        \midrule
        M2-Mixer & $46.66 \pm 0.44$ & $16.7$ & $73.20 \pm 0.2$  & $8.3$ & $78.32 \pm 0.3$ & $0.029$ \\
        \midrule
        MixMAS & $\mathbf{49.58 \pm 0.5}$ & $10.37$ & $\mathbf{75.79 \pm 0.3}$ & $9.33$ & $78.3 \pm 0.73$ & $0.033$ \\
        \bottomrule
    \end{tabular}
\end{table*}

\subsection {Experimental Setup}\label{subsec:datasets}

In this section, we describe the datasets we used in the experiments and the experimental setup.

\textbf{MM-IMDB} \footnote{https://github.com/johnarevalo/gmu-mmimdb} is a multimodal dataset with images (movie posters) and text (plots) for genre classification. We used BERT~\cite{devlin2018bert} for text embeddings. \textbf{AV-MINST} \footnote{https://github.com/slyviacassell/\_MFAS/tree/master} combines MNIST \footnote{https://yann.lecun.com/exdb/mnist/} images with FSDD \footnote{https://github.com/Jakobovski/free-spoken-digit-dataset} (digit pronunciations). \textbf{MIMIC-III} \footnote{https://physionet.org/content/mimiciii/1.4/} is a clinical dataset with time-series (12 hourly medical measurements over 24 hours) and tabular data.

For micro benchmarks, we use a learning rate of 0.001, training with sampled data from the sampler for 10 epochs. For full training, we start with a learning rate of 0.001 on MM-IMDB, using a scheduler that reduces it by a factor of 10 if validation loss shows no improvement for 2 epochs. For AV-MINST and MIMIC-III, we follow the training setup of M2-Mixer \cite{bezirganyan2023}.
For the MM-IMDB dataset, we compute the weighted F1 score due to the imbalanced labels. The other datasets were evaluated conducted using the accuracy metric.

We use MLP-Mixers, RaMLP~\cite{LaiDGZ23}, HyperMixer~\cite{mai2023hypermixer} and MonarchMixer~\cite{mai2023hypermixer} as candidates for encoder function, and HyperMixer and MLPMixer as candidates for Fusion Network. In MIMIC we opted for a fixed, simple feed-forward MLP as an encoder for the tabular modality. We compare MixMAS's performance against M2-Mixer.

We utilized two internal clusters with NVIDIA GeForce RTX 3090, GeForce RTX 2080, A40 and V100 GPUS. The total runtime of all experiments was 144 hours.

\subsection{Results}
Table \ref{tab:micro_benchmark} summarizes the micro-benchmarking results, where the pipeline selects the highest-scoring modules to construct the final architecture. The results show that there is no universal solution for modality encoders or fusion networks, as different datasets and modalities benefit from distinct modules. This underscores the strength of the MixMAS pipeline in tailoring the architecture and fusion function to the task. Notably, ConcatFusion is consistently selected during the Fusion Function stage, validating our assumption that concatenation preserves more information from the modalities compared to mean or max pooling, improving overall model performance.

Table \ref{tab:merged_res}  presents the results of training the final model on full datasets. On MM-IMDB, MixMAS surpasses M2-Mixer, achieving an average F1-weighted score of 49.58\% compared to M2-Mixer's 42.3\%, with fewer parameters (10.37 million vs. 16.7 million). For AV-MNIST, MixMAS also outperforms M2-Mixer, achieving 75.79\% average accuracy versus 73.2\%. Results for MIMIC-III show similar performance between MixMAS and M2-Mixer. We hypothesize that the tabular modality's incompatibility with MLP-Mixers, alongside fixing a simple MLP for this modality, reduces the search space.

\section{Conclusion}\label{sec:conclusion}

In this paper, we introduce MixMAS, a framework for selecting optimal MLP-based architectures using sampling and micro-benchmarking. Our approach builds on the simplicity and efficiency of MLP-Mixers, extending them to multimodal learning. Experiments confirm the framework’s effectiveness on bi-modal datasets, and future work includes testing on datasets with more modalities. We also plan to explore alternative sampling methods, like uncertainty and diversity sampling, and expand the search to include a broader range of modules.

\bibliographystyle{abbrvnat}
\bibliography{references}

\begin{thebibliography}{15}
\providecommand{\natexlab}[1]{#1}
\providecommand{\url}[1]{\texttt{#1}}
\expandafter\ifx\csname urlstyle\endcsname\relax
  \providecommand{\doi}[1]{doi: #1}\else
  \providecommand{\doi}{doi: \begingroup \urlstyle{rm}\Url}\fi

\bibitem[Baltru{\v{s}}aitis et~al.(2018)Baltru{\v{s}}aitis, Ahuja, and Morency]{baltrusaitis2017multimodal}
T.~Baltru{\v{s}}aitis, C.~Ahuja, and L.-P. Morency.
\newblock Multimodal machine learning: A survey and taxonomy.
\newblock \emph{IEEE transactions on pattern analysis and machine intelligence}, 41\penalty0 (2):\penalty0 423--443, 2018.

\bibitem[Bezirganyan et~al.(2023)Bezirganyan, Sellami, Berti-{\'E}Quille, and Fournier]{bezirganyan2023}
G.~Bezirganyan, S.~Sellami, L.~Berti-{\'E}Quille, and S.~Fournier.
\newblock M2-mixer: A multimodal mixer with multi-head loss for classification from multimodal data.
\newblock In \emph{2023 IEEE International Conference on Big Data (BigData)}, pages 1052--1058. IEEE, 2023.

\bibitem[Devlin et~al.(2019)Devlin, Chang, Lee, and Toutanova]{devlin2018bert}
J.~Devlin, M.-W. Chang, K.~Lee, and K.~Toutanova.
\newblock Bert: Pre-training of deep bidirectional transformers for language understanding.
\newblock In \emph{Proceedings of naacL-HLT}, volume~1, page~2. Minneapolis, Minnesota, 2019.

\bibitem[Fu et~al.(2023)Fu, Arora, Grogan, Johnson, Eyuboglu, Thomas, Spector, Poli, Rudra, and R{\'{e}}]{osti_10494106}
D.~Y. Fu, S.~Arora, J.~Grogan, I.~Johnson, E.~S. Eyuboglu, A.~W. Thomas, B.~Spector, M.~Poli, A.~Rudra, and C.~R{\'{e}}.
\newblock Monarch mixer: {A} simple sub-quadratic gemm-based architecture.
\newblock In A.~Oh, T.~Naumann, A.~Globerson, K.~Saenko, M.~Hardt, and S.~Levine, editors, \emph{Advances in Neural Information Processing Systems 36}, 2023.

\bibitem[Hogg et~al.(2013)Hogg, Tanis, and Zimmerman]{hogg2013probability}
R.~Hogg, E.~Tanis, and D.~Zimmerman.
\newblock \emph{Probability and Statistical Inference}.
\newblock Pearson, 2013.
\newblock ISBN 9780321923271.

\bibitem[Lai et~al.(2023)Lai, Du, Guo, and Zhang]{LaiDGZ23}
S.~Lai, X.~Du, J.~Guo, and K.~Zhang.
\newblock Ramlp: Vision mlp via region-aware mixing.
\newblock In \emph{IJCAI}, pages 999--1007, 2023.

\bibitem[Liang et~al.(2022)Liang, Zadeh, and Morency]{liang2022foundations}
P.~P. Liang, A.~Zadeh, and L.-P. Morency.
\newblock Foundations and recent trends in multimodal machine learning: Principles, challenges, and open questions.
\newblock \emph{CoRR}, abs/2209.03430, 2022.

\bibitem[Liu et~al.(2022)Liu, Li, Tao, Liang, and Zheng]{liu2022ready}
R.~Liu, Y.~Li, L.~Tao, D.~Liang, and H.~Zheng.
\newblock Are we ready for a new paradigm shift? a survey on visual deep mlp.
\newblock \emph{Patterns (N Y)}, 3\penalty0 (7):\penalty0 100520, 2022.
\newblock \doi{10.1016/j.patter.2022.100520}.

\bibitem[Mai et~al.(2023)Mai, Pannatier, Fehr, Chen, Marelli, Fleuret, and Henderson]{mai2023hypermixer}
F.~Mai, A.~Pannatier, F.~Fehr, H.~Chen, F.~Marelli, F.~Fleuret, and J.~Henderson.
\newblock Hypermixer: An mlp-based low cost alternative to transformers, 2023.

\bibitem[P{\'e}rez-R{\'u}a et~al.(2019)P{\'e}rez-R{\'u}a, Vielzeuf, Pateux, Baccouche, and Jurie]{perez2019mfas}
J.-M. P{\'e}rez-R{\'u}a, V.~Vielzeuf, S.~Pateux, M.~Baccouche, and F.~Jurie.
\newblock Mfas: Multimodal fusion architecture search.
\newblock In \emph{Proceedings of the IEEE/CVF Conference on Computer Vision and Pattern Recognition}, pages 6966--6975, 2019.

\bibitem[Stahlschmidt et~al.(2022)Stahlschmidt, Ulfenborg, and Synnergren]{stahlschmidt2022multimodal}
S.~R. Stahlschmidt, B.~Ulfenborg, and J.~Synnergren.
\newblock Multimodal deep learning for biomedical data fusion: a review.
\newblock \emph{Briefings Bioinform.}, 23\penalty0 (2), 2022.

\bibitem[Tolstikhin et~al.(2021)Tolstikhin, Houlsby, Kolesnikov, Beyer, Zhai, Unterthiner, Yung, Steiner, Keysers, Uszkoreit, Lucic, and Dosovitskiy]{tolstikhin2021}
I.~O. Tolstikhin, N.~Houlsby, A.~Kolesnikov, L.~Beyer, X.~Zhai, T.~Unterthiner, J.~Yung, A.~Steiner, D.~Keysers, J.~Uszkoreit, M.~Lucic, and A.~Dosovitskiy.
\newblock Mlp-mixer: An all-mlp architecture for vision.
\newblock In \emph{Advances in Neural Information Processing Systems 34}, pages 24261--24272, 2021.

\bibitem[Xu et~al.(2021)Xu, So, and Dai]{xu2021mufasa}
Z.~Xu, D.~R. So, and A.~M. Dai.
\newblock Mufasa: Multimodal fusion architecture search for electronic health records.
\newblock In \emph{Proceedings of the AAAI Conference on Artificial Intelligence}, volume~35, pages 10532--10540, 2021.

\bibitem[Yin et~al.(2022)Yin, Huang, and Zhang]{yin2022bm}
Y.~Yin, S.~Huang, and X.~Zhang.
\newblock Bm-nas: Bilevel multimodal neural architecture search.
\newblock In \emph{Proceedings of the AAAI Conference on Artificial Intelligence}, volume~36, pages 8901--8909, 2022.

\bibitem[Yu et~al.(2020)Yu, Cui, Yu, Wang, Tao, and Tian]{yu2020deep}
Z.~Yu, Y.~Cui, J.~Yu, M.~Wang, D.~Tao, and Q.~Tian.
\newblock Deep multimodal neural architecture search.
\newblock In \emph{Proceedings of the 28th ACM International Conference on Multimedia}, pages 3743--3752, 2020.

\end{thebibliography}

\end{document}